\documentclass[10pt,twocolumn,letterpaper]{article}

\usepackage{cvpr}              

\definecolor{cvprblue}{rgb}{0.21,0.49,0.74}
\definecolor{yujieColor}{rgb}{0,0.4,0}

\newcommand{\methodName}{CoSMo3D}

\usepackage{booktabs}
\usepackage{makecell} 
\usepackage[table]{xcolor} 
\usepackage{adjustbox}
\usepackage{amsthm}
\usepackage{amsmath}    
\usepackage{graphicx}   
\usepackage{multirow}   
\usepackage{booktabs}   
\usepackage{mathtools}
\usepackage{threeparttable}
\usepackage[pagebackref,breaklinks,colorlinks,allcolors=cvprblue]{hyperref}

\def\paperID{1784} 
\def\confName{CVPR}
\def\confYear{2026}

\title{CoSMo3D: Open-World Promptable 3D Semantic Part Segmentation through LLM-Guided Canonical Spatial Modeling}

\author{
\textbf{Li Jin$^{1*}$, Weikai Chen$^{2*}$, Yujie Wang$^{3\dagger}$, Yingda Yin$^2$, Zeyu HU$^2$, Runze Zhang$^2$\vspace{0.07cm}}\\
\textbf{Keyang Luo$^2$, Shengju Qian$^2$, Xin Wang$^2$, Xueying Qin$^{1\dagger}$\vspace{0.35cm}} \\
 \\
$^1$SDU \quad $^2$LIGHTSPEED \quad $^3$UNC Chapel Hill
}

\begin{document}


\twocolumn[{%
\renewcommand\twocolumn[1][]{#1}%
\maketitle
\begin{center}
    \centering
    \vspace{-3mm}
    \includegraphics[width=\linewidth]{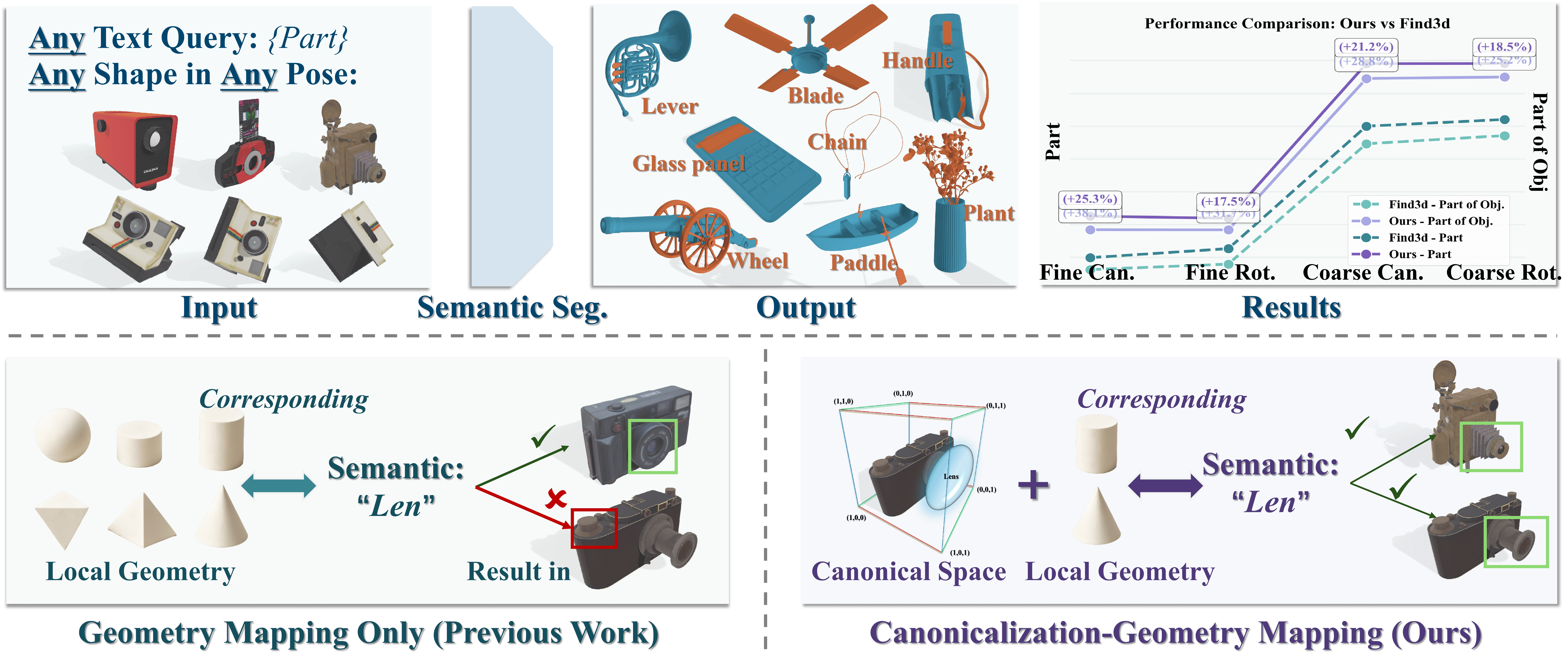}
    \vspace{-7mm}
    \captionof{figure}{We propose CoSMo3D, an open-world promptable 3D semantic segmentation method. It introduces canonical space perception to break the limitation of any pose and shape, achieves state-of-the-art performance across multiple settings, and significantly outperforms geometry-mapping-only methods.}
    \label{fig:schematic}
    \vspace{-1mm}
\end{center}
}]

\def\thefootnote{*}\footnotetext{Equal Contribution.}
\def\thefootnote{$\dagger$}\footnotetext{Corresponding Author.}
\setcounter{footnote}{0} 


\begin{abstract}
Open-world promptable 3D semantic segmentation remains brittle as semantics are inferred in the input sensor coordinates. Yet, humans, in contrast, interpret parts via functional roles in a canonical space -- wings extend laterally, handles protrude to the side, and legs support from below. Psychophysical evidence shows that we mentally rotate objects into canonical frames to reveal these roles. To fill this  gap, we propose \methodName{}, which attains canonical space perception by inducing a latent canonical reference frame  learned directly from data.  By construction, we create a unified canonical dataset through LLM-guided intra- and cross-category alignment, exposing canonical spatial regularities across 200 categories. 
By induction, we realize canonicality inside the model through a dual-branch architecture with canonical map anchoring and canonical box calibration, collapsing pose variation and symmetry into a stable canonical embedding. This shift from input pose space to canonical embedding yields far more stable and transferable part semantics. Experimental results show that \methodName{} establishes new state of the art in open-world promptable 3D segmentation. Project page: \url{https://github.com/JinLi998/CoSMo3D/tree/main}
\end{abstract}


\section{Introduction}
\label{sec:intro}

Humans perceive 3D objects by reasoning jointly over geometry and spatial semantics. When we recognize ``legs of chair", we rely not only on their slender shape but also on their \textit{canonical position} -- below the seat and supporting the object. Psychophysical studies~\cite{Roger1971mental} further suggest that humans perform \textit{mental rotation} to align observed objects with canonical poses, allowing us to identify parts consistently across poses and categories. 
This \textit{canonical space reasoning} is a core component of human 3D understanding, yet remains largely absent in current segmentation models.

Recent advances in open-world 3D understanding have started to bridge this gap. In particular, Find3D~\cite{ma2025find} introduced the task of open-world promptable 3D segmentation, where a model segments any 3D object according to a free-form text query (e.g. ``handle", ``wing", or ``paddle"). This is an important step toward scalable 3D understanding: rather than limiting segmentation to fixed categories, a model can interpret open-ended prompts and transfer knowledge to unseen categories.
Find3D achieves this by learning direct alignment between geometry features and language embeddings, supported by a large-scale automatically labeled dataset.
It makes a crucial step toward general 3D perception, demonstrating impressive zero-shot generalization and flexible promptability.

Despite its success, Find3D and similar methods remain fundamentally limited by their reliance on geometric-text matching.
They assume that geometrically similar shapes tend to share similar semantics, yet this correlation often breaks in practice.
For instance, the arms and legs of a chair may appear geometrically similar but serve distinct functions, while airplane wings and bird wings differ in form yet share the same semantics.
Without explicit reasoning grounded in canonical-space regularities, these models lack awareness of where a semantic part should occur relative to the whole object, leading to inconsistent predictions under pose variations, symmetries, or across categories.
For instance, in our teaser example (\Cref{fig:schematic}), Find3D misclassifies a cylindrical. 
Although data augmentation provides limited robustness, these models still lack an internal notion of \textit{spatial semantics}, a core ingredient of human perception.


We argue that achieving human-like 3D understanding requires canonical space perception: the ability to internalize a canonical reference space shared across shapes and categories, and interpret part semantics w.r.t. this \textit{reference} —- not the raw input pose. 
To this end, we introduce \textit{\methodName{}}, an LLM-guided open-world promptable 3D segmentation framework that achieves \textbf{canonical space perception} by \textbf{inducing a {latent} reference frame} from data.
This principle manifests in two levels in \methodName{}.
1) \textit{Externally}, we construct a unified canonical dataset through an LLM-guided intra- and cross-category canonicalization pipeline, so that canonicality is no longer defined only within a category but is shared across object families. This gives us a common \textit{substrate} that generalized beyond category boundaries.
2) \textit{Internally}, we instantiate canonicaliity by inducing a latent canonical reference frame directly from data, instead of prescribing canonical poses manually.
Within this framework, we propose a dual-branch architecture: the primary branch retains Find3D's geometry-language alignment, while a canonical branch \textit{pulls} point embeddings toward canonical codes and \textit{tightens} part-level spatial extents via two canonical-space objectives (canonical map anchoring and canonical box calibration). 
Canonical space thus becomes an emergent attractor manifold: different poses, symmetries, and shape variants of the same functional part collapse to a single canonical embedding, making semantics spatially grounded and inherently pose-invariant.

Extensive experiments demonstrate that \methodName{} substantially outperforms existing methods. By integrating language-driven semantics, geometry understanding, and canonical spatial reasoning under a unified framework, \methodName{} bridges the gap between geometry-text alignment and human-like spatial perception. 
Overall, our contributions can be summarized as:

\begin{itemize}
    \item We reframe open-world 3D segmentation as reasoning grounded in canonical-space regularities, rather than matching geometry to text in input pose.
    \item We make canonicality a learnable latent structure, instantiated through an LLM-aligned canonical dataset and dual-branch canonical regularization.
    \item \methodName{} sets the new state of the art in open-world 3D segmentation in terms of accuracy, spatial stability, and cross-category generalization.
\end{itemize}

\begin{figure*}[htbp]
\centering
\includegraphics[width=1\linewidth]{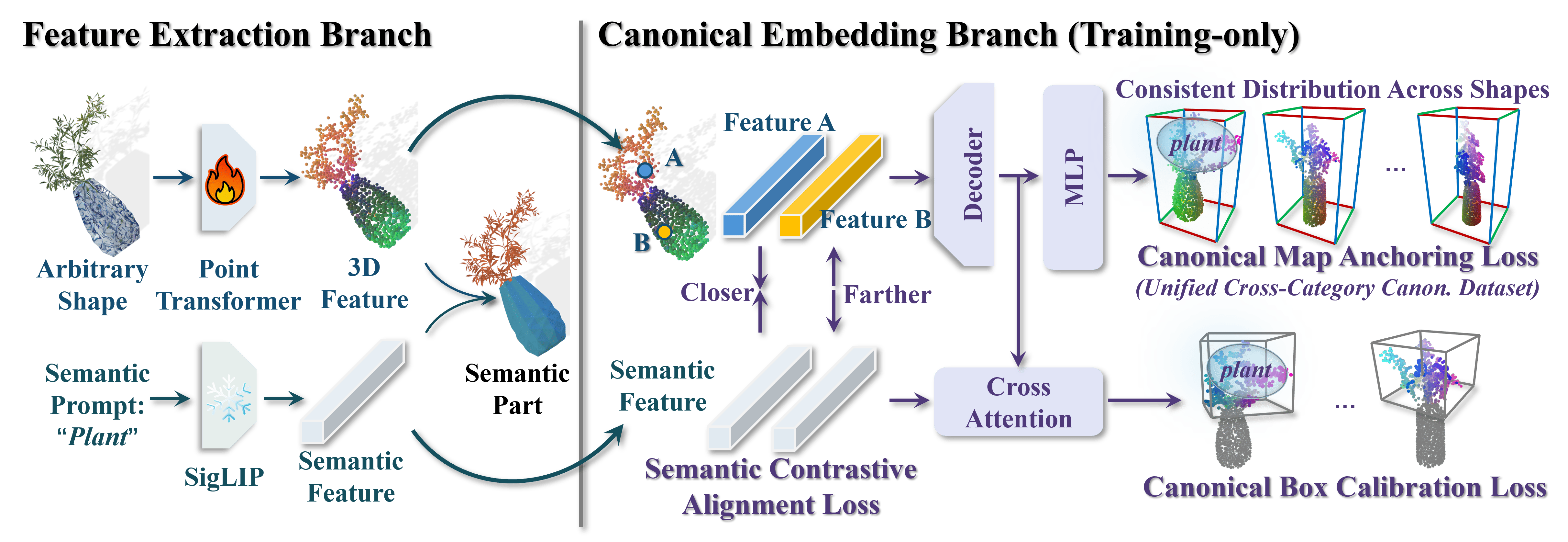}
    \vspace{-6mm}
  \caption{We propose a dual-branch framework for open-world promptable 3D semantic segmentation: the feature extraction branch encodes 3D shape features (via Point Transformer) and text semantic features (via SigLIP) to enable cross-modal part segmentation. A training-only canonical embedding branch then enforces consistent canonical space perception via semantic contrastive alignment, canonical map anchoring, and canonical box calibration losses, ensuring robust reasoning across any shape in any pose.}
  \label{fig:framework}
  \vspace{-3mm}
\end{figure*}

\section{Related Work}
\label{sec:related}

Traditional data-driven part segmentation~\cite{zhang2025generative,ma2022rethinking,zhao2021point,girshick2015region} takes 2D images or 3D shapes as input. Its core lies in predefining object categories and part labels, followed by pixel-wise or point-wise classification. However, due to its heavy reliance on predefined labels, it has limited generalization ability and cannot adapt to open-world scenarios. In the context of open-world scenarios, {2D segmentation technology} has achieved breakthroughs first. Among them, {class-agnostic segmentation} represented by SAM~\cite{kirillov2023segment,ravi2024sam} does not require predefined categories, balancing segmentation accuracy and generalization. SAM3 further introduces {prompt semantic segmentation}, which improves alignment with human semantic cognition and controllability through the mapping between open vocabulary and parts, providing a reference for the development of 3D segmentation techniques.

\textbf{Class-Agnostic 3D Segmentation} takes 3D shapes as input, with the goal of realizing multi-level decomposition of objects based on geometric features without predefined categories. It mainly includes two technical paths: One is 2D conversion-based methods~\cite{yang2023sam3d,tang2024segment}, which convert 3D shapes into 2D images through rendering, use 2D models such as SAM for segmentation, then back-project the results to 3D space, and finally alleviate issues like multi-view consistency and self-occlusion through post-processing. However, this path relies on conversion and post-processing, resulting in slow speed and unavoidable multi-view consistency problems. The other is pure 3D methods~\cite{yang2024sampart3d,yan2025x,zhu2025partsam,deng2025geosam2,liu2025partfield}: Sampart3D~\cite{yang2024sampart3d} implements segmentation through a 3D feature extractor combined with test-time adaptation (TTA), but TTA increases computational costs; Partfield~\cite{liu2025partfield} and its subsequent improved methods~\cite{yan2025x,zhu2025partsam,deng2025geosam2} expand the feature differences between parts through contrastive learning and then complete segmentation via clustering. Although they achieve better performance with stronger network architectures and more data, they are all limited to low-level part segmentation, lacking high-level semantic understanding, with poor interactivity and controllability. Additionally, the object features do not have consistency under arbitrary poses, making it difficult to establish a mapping with semantics.

\textbf{Promptable 3D Semantic Segmentation} takes 3D shapes and semantic text as input, with the core of establishing a mapping between semantics and 3D parts through text conditions to output target segmentation results. It is divided into two types of methods: One is 2D-based methods~\cite{abdelreheem2023satr,umam2024partdistill,zhu2023pointclip,zhou2023partslip++,liu2023partslip,takmaz2023openmask3d}, which adopt 2D semantic segmentation models such as Glip~\cite{radford2021learning} and focus on solving the consistency problem when aggregating 2D results into 3D space. However, relying on 2D segmentation, they can only handle objects in upright poses (e.g., it fails when objects are upside down), and the conversion process significantly increases time costs. The other is pure 3D methods~\cite{ma2025find}, which train semantic Prompt models based on 3D datasets, supporting segmentation for ``arbitrary text, arbitrary objects, and arbitrary poses" with strong robustness and no need for 2D conversion, thus achieving a significant speed improvement. However, after breaking the constraints of the canonical space, the network only learns the correspondence between ``text and geometric structures" through the constraint of ``text-part similarity". Although similar semantic parts often have similar geometric structures (e.g., the legs of a chair are mostly slender), similar geometric structures may correspond to different semantics (e.g., the arm and leg of a chair are both slender but belong to different semantics). Therefore, additional perceptual capabilities are still needed to address semantic ambiguity in open-world segmentation.
\section{Method}

\subsection{Overview}
Given a 3D shape and a text prompt, our method performs promptable semantic 3D segmentation by encoding geometric features from the shape and semantic features from the text, computing cross-modal similarity to associate text with shape regions, and decoding these associations into part-level labels. An overview is given in \Cref{fig:framework}.

To equip the framework with a {transferable} understanding of canonical space across categories for open-world 3D semantic part reasoning, we operationalize this goal through both \emph{external} and \emph{internal} fronts.
\textbf{Data \& Supervision (External)} (\Cref{sec:data_create}).
We construct a cross-category canonical dataset using an LLM-guided {intra- and cross-category} canonicalization pipeline. This corpus provides supervisory signals, including canonical maps, part boxes, and semantic associations, that enable the model to {induce a latent canonical reference frame} from data.
\textbf{Dual-Branch Framework \& Canonical-Space Objectives (Internal)} (\Cref{sec:arch}).
As \Cref{fig:framework} shows, our dual-branch architecture consists of: 
(i) a feature-extraction branch used during both training and inference, and 
(ii) a training-only canonical branch that predicts canonical maps and part-level boxes. 
Our training loss combines a {geometry–text alignment} term that stabilizes cross-modal mapping with two canonical-space regularizers, i.e., {canonical-map anchoring} and {canonical-box calibration}, which enforce consistency with the induced reference frame by pulling point embeddings toward canonical codes and tightening the spatial extent of semantic parts in canonical space.


\subsection{Unified Cross-Category Canonical Dataset} \label{sec:data_create}

\begin{figure}[t]
\centering
  \includegraphics[width=0.99\linewidth]{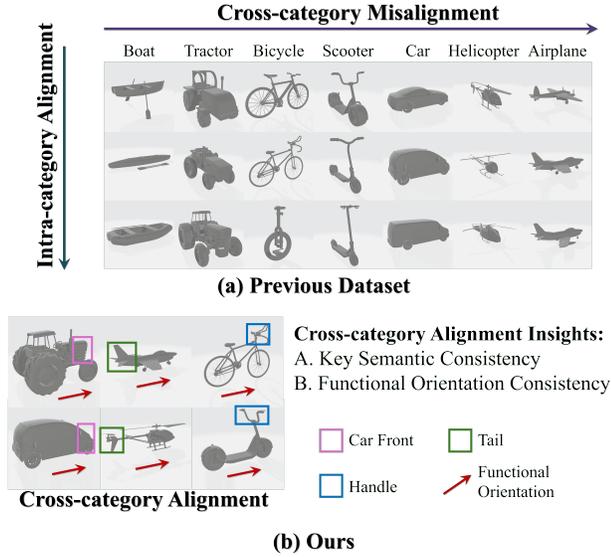} 
    \vspace{-2mm}
  \caption{\textbf{(a)} Prior works perform category-level canonicalization, aligning intra-category shapes but neglecting cross-category consistency. \textbf{(b)} We cluster categories via LLM and align different categories relying on key semantic parts and functional consistency.}
  \label{fig:dataset}
  \vspace{-0.55cm}
\end{figure}

Learning transferable canonical space priors is essential for generalizing to diverse and potentially unseen categories in open-world settings. However, existing canonical datasets~\cite{wang2019normalized,chang2015shapenet} treat categories individually, without cross-category alignment (as shown in Fig.\ref{fig:dataset} (a) ), limiting scalability to open-world scenarios. To address this, we construct a \emph{unified canonical dataset} that covers most common object categories with cross-category alignment (Fig.\ref{fig:dataset} (b)). The process involves two key steps: (i) \emph{intra-category canonicalization}, which aligns instances within each category to a shared canonical space; and (ii) \emph{cross-category canonicalization}, which aligns semantically corresponding parts or landmarks across categories.


As intra-category canonicalization has been extensively studied in existing work~\cite{jin2025one,wu2023omniobject3d,ahmed20243dcompat200}, we focus primarily on the more challenging task of cross-category canonicalization. This is difficult due to large variations in geometry and function across categories. For example, categories like forks, bicycles, and tree trunks differ significantly in both shape and usage. To tackle this, we design a hierarchical alignment strategy: we first group object categories into semantically coherent \emph{category clusters} based on shared functions or usage contexts; we then perform intra-cluster and cross-cluster alignment in sequence, gradually building a unified canonical space across all categories.


In detail, we build a category-level canonical dataset~\cite{ahmed20243dcompat200} comprising 200 common categories and 17K shapes. First, we employ a large language model (e.g., GPT) to cluster the 200 categories into 19 semantically coherent groups (e.g., transportation, tools). Within each cluster, alignment is performed based on shared functional characteristics. For instance, steering-related parts of bicycles and airplanes within the transportation cluster are aligned to ensure a consistent orientation. Since the dataset~\cite{ahmed20243dcompat200} already unifies the core semantic direction within each category, aligning multiple categories within the same cluster typically requires only simple discrete rotations (e.g., 90°, 180°, or 270°), making the process computationally efficient. Next, cross-cluster alignment is performed by verifying high-level semantic consistency. For example, we ensure that the transportation and animal clusters share a consistent forward movement direction. Moreover, to further enrich shape diversity, we apply axis-aligned deformations to objects. Ultimately, we obtain a unified canonical dataset with consistent alignment across and within categories. We provide additional dataset details and sample visualizations in the Supplementary Material.

\subsection{Canonical‑Aware Dual‑Branch Framework} \label{sec:arch}
\noindent \textbf{Architecture.} We design a two-branch framework for promptable 3D semantic understanding. As \Cref{fig:framework} shows, we adopt a {feature extraction branch} to predict 3D features from the input point cloud in a \emph{single inference pass}, in contrast to prior 3D aggregation methods that rely on 2D renderings. The {canonical embedding branch}, introduced during training as a unique part of our design, generates intermediate representations supervised by canonical-space-related signals. This enhances the model’s perception of canonical space, improving alignment across shapes and prompts.
The feature extraction branch largely follows the design of Find3D. It employs Pt3~\cite{wu2024point} as the backbone for point cloud encoding and SigLIP~\cite{zhai2023sigmoid} for text feature extraction. A lightweight 3-layer MLP projects point features into the same embedding space as the text features.

The canonical embedding branch contains two heads: one for \emph{canonical map prediction}, and the other for \emph{semantic bounding box prediction}. For canonical map prediction, we use 3D shape features as input and, inspired by 3D generation methods~\cite{li2024craftsman3d,hunyuan3d2025hunyuan3d}, regress three continuous scalar fields (encoded as RGB color maps) rather than discrete per-point values, to better preserve spatial continuity. For semantic bounding box prediction, text features serve as queries to extract relevant regions from the shape features, which are output as a 6-dimensional vector representing the bbox.

\begin{figure}[t]
\centering
  \includegraphics[width=0.99\linewidth]{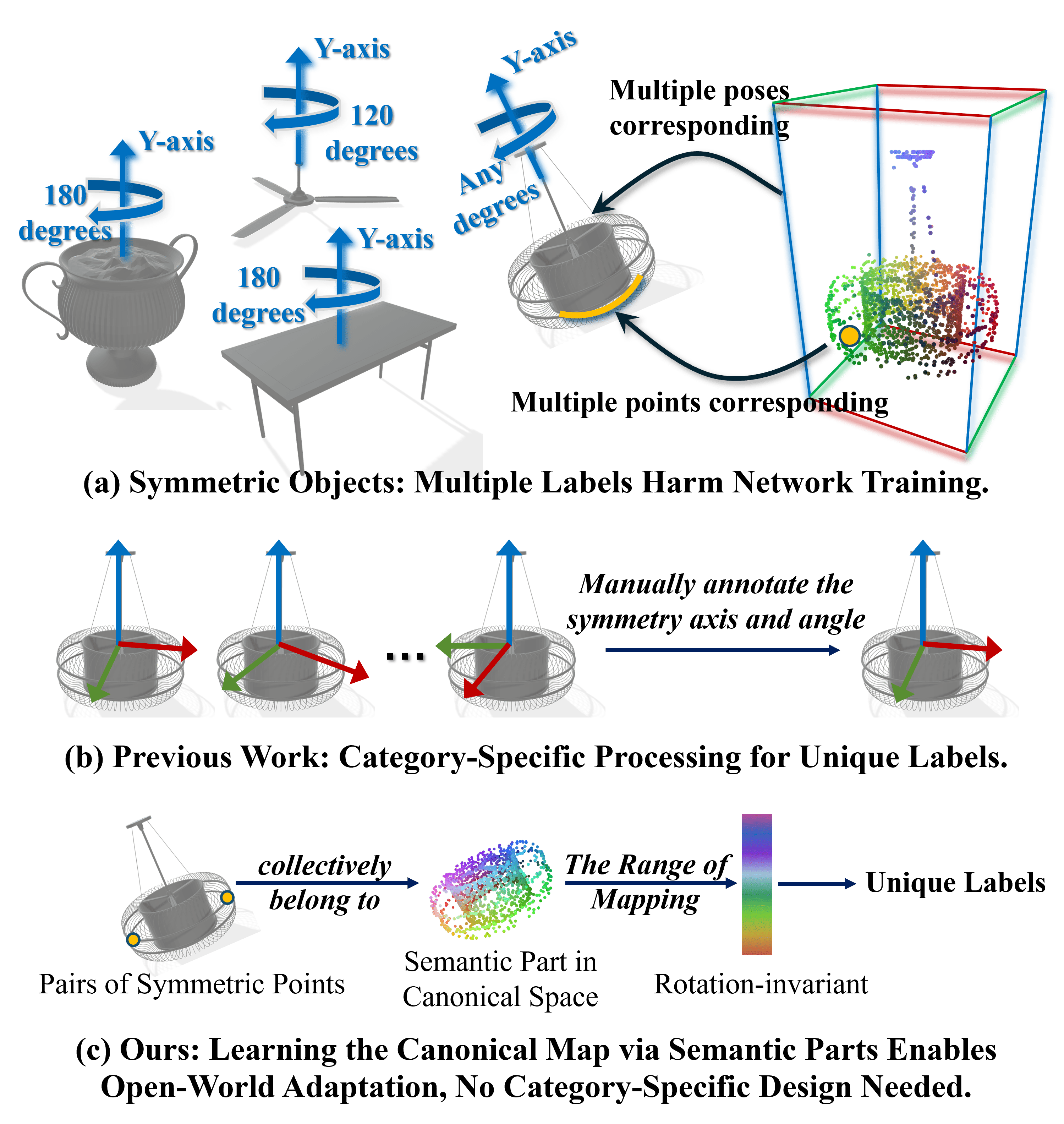} 
    \vspace{-1mm}
  \caption{Handling Symmetric Objects for Canonical Map Anchoring.
(a) For symmetric shapes, multiple valid poses induce ambiguous point-wise canonical labels, making direct point-wise supervision unreliable. 
(b) Prior works rely on category-specific processing with manual symmetry annotation, limiting open-world scalability.
(c) We instead apply an order-invariant set loss on RGB-encoded canonical coordinates that matches the overall layout of semantic parts in canonical space while remaining robust to symmetric pose ambiguities.}
  \label{fig:perception}
  \vspace{-0.55cm}
\end{figure}

\subsubsection{Semantic Contrastive Alignment Loss} \label{sec:contrastive}

In open-world promptable part segmentation, semantics are non-exclusive: a single part may be referred to by material, function, synonym choice, or label granularity. A point on a bottle’s neck can validly be “neck” (fine) or “body” (coarse), making single-point single-label supervision brittle. Following Find3D~\cite{ma2025find}, we can adopt conduct contrastive learning to encode soft point–text affinities:
\vspace{-2mm}
\begin{equation}
\mathcal{L}_h =  \frac{1}{M} \sum_{i=1}^M - \log\frac{f(\bar{\mathbf{p}}_i,\mathbf{t}_i)}{\sum_{k\in\mathcal{B}}f(\bar{\mathbf{p}}_i,\mathbf{t}_k)},
\label{eq:contr}
\vspace{-2mm}
\end{equation}

\noindent where $\bar{\mathbf{p}}_i$ denotes the average feature of point samples within the mask of the $i$-th semantic part, $\mathbf{t}_i$ is the text embedding. $M$ is the number of part labels and $\mathcal{B}$ denotes the set of semantic labels in the minibatch, and $f(\bar{\mathbf{p}}_i,\mathbf{t}_i)=\exp(\bar{\mathbf{p}}_i^\top \cdot \mathbf{t}_i/\tau)$ with temperature $\tau$. 
We sample points in each part strategically, as detailed below.

We observe that uniform point sampling within each part biases Eq.(\ref{eq:contr}) toward block--text consistency while possibly ignoring deviations of individual points, yielding noisy masks near part boundaries and slower convergence. To address this, we introduce a Hard Negative Sampling strategy, i.e., sampling more discriminative negative samples from inter-part boundaries to improve segmentation accuracy. The hard bidirectional contrastive loss is defined as:
\begin{small}
{
\begin{alignat}{2}
   \mathcal{L}_h &= \frac{1}{2M}\sum_{i=1}^M - \log\frac{f(\bar{\mathbf{p}}_i,\mathbf{t}_i)}{\sum_{k\in \mathcal{B}_i}f(\bar{\mathbf{p}}_i, \mathbf{t}_k)} -\log\frac{f(\mathbf{t}_i,\bar{\mathbf{p}}_i)}{\sum_{n\in \mathcal{P}_i}f(\mathbf{t}_n,\hat{\mathbf{p}}_n)}, & \nonumber \\
   \hat{\mathbf{p}}_n &=\frac{1}{W_n}\left( \textstyle \sum_{\mathbf{p}_j\in \Omega_n} \mathbf{p}_j + (1+\alpha) \textstyle \sum_{\mathbf{p}_e \in \mathcal{E}_n} \mathbf{p}_e\right), & \label{eq:contr}
\end{alignat}
}
\end{small}

\noindent where $\alpha = 0$ for $n = i$, and $\alpha > 0$ for $n \ne i$. 
$\mathbf{p}_j$ and $\mathbf{p}_e$ denote individual point features. 
$\Omega_n$ and $\mathcal{E}_n$ represent the interior and edge regions of the part associated with semantic label $n$, respectively. 
The normalization weight is defined as
$W_n = |\Omega_n| + (1 + \alpha)|\mathcal{E}_n|$.

\subsubsection{Canonical Space Regularization} \label{sec:canon_reg}
Built on the dual-branch design and the curated cross-category canonical dataset, our method regularizes point features with canonical space cues. We introduce two losses, detailed below, to align per-part canonical distributions and enforce robust spatial extents.

\noindent \textbf{Canonical Map Anchoring Loss.} 
Our goal is to enforce that semantic parts with the same (or closely related) semantics exhibit consistent spatial distributions in canonical space, both within and across object categories. We achieve this by anchoring per-part canonical maps using the canonicalization metadata and semantic part labels provided in our curated dataset. 
However, in practice, we observe that axial and planar symmetries often introduce correspondence ambiguities (Figure~\ref{fig:perception} (a)): multiple poses of a symmetric object are equally valid in canonical space, making point-wise canonical supervision unreliable. 
Existing canonical space perception methods attempt to resolve this by manually annotating symmetry axes or enforcing category-specific constraints (Figure~\ref{fig:perception} (b)), but such designs do not scale to open-world segmentation.


Our key insight is to \textit{avoid correspondence altogether}.
Instead of supervising canonical coordinate point-wisely, we treat each semantic part as a distribution in canonical space.
We then regularize the predicted canonical map by matching it to the ground-truth canonical distribution using a bidirectional Chamfer distance.
As the objective compares \textit{shapes of distributions} rather than individual coordinates, symmetric configurations become equivalent in canonical space. Symmetric points therefore converge to the same canonical region automatically, sidestepping the need for symmetry labels or per-category axis annotations.

Formally, let $\mathcal{G}^{p}_{m}=\{\mathbf{a}_i\}$ and $\mathcal{G}^{t}_{m}=\{\mathbf{b}_j\}$ denote the predicted and ground-truth point sets in the canonical space $\Omega\subset\mathbb{R}^d$ (typically $d\in\{2,3\}$) for part $m$. We define the Canonical Map Anchoring loss as:

\begin{small}
\begin{alignat}{2}
 \mathcal{L}_{ca}
= \frac{1}{M}  \textstyle \sum_{m=1}^{M} & \Big( 
\frac{1}{\lvert \mathcal{G}^{p}_{m}\rvert} \textstyle \sum_{\mathbf{a}_i\in \mathcal{G}^{p}_{m}} \min_{\mathbf{b}_j\in \mathcal{G}^{t}_{m}} \|\mathbf{a}_i-\mathbf{b}_j\|_{p}
+   \nonumber \\
&\frac{1}{\lvert \mathcal{G}^{t}_{m}\rvert} \textstyle \sum_{\mathbf{b}_j\in \mathcal{G}^{t}_{m}} \min_{\mathbf{a}_i\in \mathcal{G}^{p}_{m}} \|\mathbf{b}_j-\mathbf{a}_i\|_{p}
\Big).   \label{eq:l_ca}
\end{alignat}
\end{small}
\!$\|\cdot\|_{p}$ denotes the $p$-norm (we use $p{=}2$ by default). 
By aligning distributions instead of points, the loss drives rotation-invariant, symmetry-robust canonical layouts.

\begin{figure*}[htbp]
\centering
\includegraphics[width=.95\linewidth]{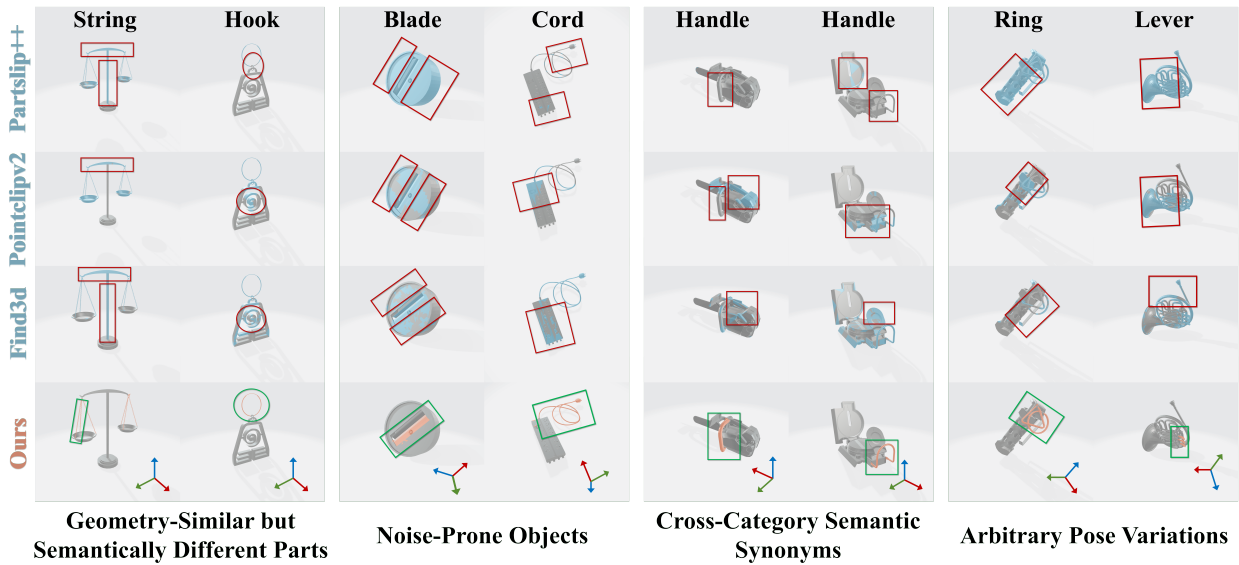}
\vspace{-4mm}
\caption{Qualitative comparison of promptable 3D part segmentation.  
Across challenging cases (similar geometry with different semantics, noise-prone objects, cross-category semantics, and arbitrary poses), our method produces more accurate and consistent part localizations than existing baselines.
}
\label{fig:comparison}
\vspace{-5mm}
\end{figure*}

\begin{table*}[t]
\centering
\caption{Quantitative evaluation for promptable semantic segmentation (with semantic labels as reference and mean IoU reported). Top: coarse-grained dataset and fine-grained dataset; Bottom: ShapeNet-Part dataset and PartNet-E dataset. Find3D$^{*}$ denotes retraining using the 3D data we constructed; PartSLIP++$^{*}$ denotes fine-tuning on the PartNet-E dataset with canonical views as input.  }
\label{tab:promptable_seg_combined}
\resizebox{\textwidth}{!}{%
\scriptsize
\begin{tabular}{c|c|cccc|cccc} 
\toprule
mIoU (\%) & \textbf{3D feedforward} & \multicolumn{4}{c|}{\textbf{3Dcompat-Coarse}} & \multicolumn{4}{c}{\textbf{3DCompat-Fine}} \\ 
\midrule
& & \multicolumn{2}{c}{Canonical} & \multicolumn{2}{c|}{Rotated} & \multicolumn{2}{c}{Canonical} & \multicolumn{2}{c}{Rotated} \\ 
& & \{Part\} of \{Obj.\} & \{Part\} & \{Part\} of \{Obj.\} & \{Part\} & \{Part\} of \{Obj.\} & \{Part\} & \{Part\} of \{Obj.\} & \{Part\}  \\
\midrule
PointCLIPV2    & $\times$ & 14.09 & 14.16 & 13.17 & 13.39 & 7.25 & 7.18 & 7.43 & 7.20 \\
PartSLIP++    & $\times$ & 6.12 & 17.62 & 5.89 & 15.82 & 3.79 & 5.68 & 3.41 & 4.76 \\
Find3D & \checkmark & 31.72 & 25.36 & 32.47 & 26.23 & 12.69 & 12.57 & 12.73 & 12.79 \\
Find3D$^{*}$ & \checkmark & 36.89 & 45.00 & 38.14 & 46.02 & 17.72 & 24.97 & 18.57 & 26.35 \\
Ours   & \checkmark & \textbf{47.51} & \textbf{54.52} & \textbf{47.74} & \textbf{54.55} & \textbf{24.48} & \textbf{31.29} & \textbf{24.46} & \textbf{30.97} \\
\midrule 
mIoU (\%) & \textbf{3D feedforward} & \multicolumn{4}{c|}{\textbf{ShapeNet-Part}} & \multicolumn{4}{c}{\textbf{PartNet-E}} \\ 
\midrule
& & \multicolumn{2}{c}{Canonical} & \multicolumn{2}{c|}{Rotated} & \multicolumn{2}{c}{Canonical} & \multicolumn{2}{c}{Rotated} \\ 
& & \{Part\} of \{Obj.\} & \{Part\} & \{Part\} of \{Obj.\} & \{Part\} & \{Part\} of \{Obj.\} & \{Part\} & \{Part\} of \{Obj.\} & \{Part\}  \\
\midrule
OpenMask3D    & $\times$ & 8.94 & 10.37 & 6.75 & 14.56 & 12.54 & 11.24 & 11.93 & 11.67 \\
PartSLIP++$^{*}$   & $\times$ & 1.43  & 6.46  & 0.94 & 6.03 & 5.12  & \textbf{32.71}  & 3.87 & \textbf{23.03} \\
PointCLIPV2    & $\times$ & 16.91  & 20.22  & 16.88 & 18.19 & 11.28 & 9.70 & 10.32 & 10.22 \\
Find3D  & \checkmark & 28.39 & 24.09 & 29.64 & 23.71 & 16.86 & 16.38 & 17.62 & 17.16 \\ 
Ours   & \checkmark & \textbf{36.16} & \textbf{33.31} & \textbf{34.20} & \textbf{32.84} & \textbf{17.59} & 17.19 & \textbf{18.48} & 18.17 \\
\bottomrule
\end{tabular}
}
\vspace{-3mm}
\end{table*}
\noindent\textbf{Canonical Box Calibration Loss.}
At inference, parts are segmented by matching point-wise features to text embeddings from user prompts.
While flexible, this procedure is susceptible to local noise and may yield fuzzy or irregular boundaries, partly because the preceding losses emphasize part-level distribution alignment over point-wise accuracy. 
To reinforce boundary consistency, the canonical branch predicts a 3D bounding box for each semantic part in canonical space.
These boxes provide a coarse but stable spatial prior that sharpens boundaries and suppress spurious activations (see Figure \ref{fig:comparison} (middle)).
The loss is formulated as:
\begin{equation}
\mathcal{L}_{\text{cb}}
= \frac{1}{M} \sum_{m=1}^{M} \frac{1}{6}
\left\| \mathbf{b}_{m}^{\mathrm{p}} - \mathbf{b}_{m}^{\mathrm{t}} \right\|_{1},
\quad \mathbf{b}_{m}^{(\cdot)} \in \mathbb{R}^{6}.
\end{equation}
Here $M$ denotes the number of semantic parts. 
For part $m$, the predicted and ground-truth boxes are 
$\mathbf{b}_{m}^{\mathrm{p}}, \mathbf{b}_{m}^{\mathrm{t}} \in \mathbb{R}^{6}$, 
parameterized as $[x_{\min},y_{\min},z_{\min},x_{\max},y_{\max},z_{\max}]$. 
This loss encourages parts to occupy coherent spatial extents in canonical space, complementing distribution-level anchoring with tighter geometric regularization.


\subsubsection{Full Training Objective}

The full training objective is formulated as:
\begin{equation}
\mathcal{L}_{\text{total}} = \lambda_h \cdot \mathcal{L}_h + \lambda_{ca} \cdot \mathcal{L}_{ca} + \lambda_{cb} \cdot \mathcal{L}_{cb},
\end{equation}
where $\lambda_h = 1$, $\lambda_{ca} = 10$, and $\lambda_{cb} = 3$ denote the balancing weights. For stable convergence, we adopt a two-stage training scheme: in stage-1, we first train the framework with the alignment loss $\mathcal{L}_h$ alone until convergence. Then we incorporate $\mathcal{L}_{ca}$ (canonical map anchoring loss) and $\mathcal{L}_{cb}$ (canonical box calibration loss) in stage-2 and continue training until convergence.

\section{Experiment}

\subsection{Implementation Details}
\textbf{Training Datasets.} 
We construct our unified canonical dataset on top of the 3Dcompat200 corpus~\cite{ahmed20243dcompat200}, which provides \(\sim\)17K shapes across 200 categories with semantic part annotations. This covers over everyday object classes supports strong generalization. For training our framework, we normalize each object to a unit bounding box and uniformly sample 5{,}000 surface points per object, retaining both RGB color and normal vectors.

\noindent
\textbf{Framework.} We adopt PointTransformerV3~\cite{wu2024point} as the backbone to extract $768$-D pointwise features, and use SigLIP-Base/16-224~\cite{zhai2023sigmoid} to obtain $768$-D text embeddings. More implementation details are provided in the supplementary material.

\subsection{Comparisons}
For the comparison of promptable semantic segmentation, we design evaluations covering data distribution, input states, and query forms, with details as follows:


\noindent\textbf{Evaluation Datasets.} We evaluate our method and representative state-of-the-art baselines on several public 3D part segmentation benchmarks. We first report results on the 3Dcompat-Coarse and 3Dcompat-Fine test sets from the 3Dcompat200 dataset~\cite{ahmed20243dcompat200}, which comprises $200$ object categories and about $2,000$ shapes annotated with part labels at coarse and fine granularities, respectively. We further assess performance on two commonly used benchmarks, ShapeNet-Part~\cite{yi2016scalable} and PartNet-E~\cite{mo2019partnet}, containing 16 and 45 object categories, respectively.


\noindent
\textbf{Test Settings.} To comprehensively evaluate the proposed method under open-world conditions, we consider two key factors commonly encountered in practice: prompt variation and object pose uncertainty. To assess robustness to prompt ambiguity, we design two input modes for textual queries: ``\{Part\}” denotes single-word prompts (e.g., \textit{leg}), while “\{part\} of a \{category\}” denotes compositional phrase prompts (e.g., \textit{leg of a chair}). To simulate pose variation in open-world scenarios, we evaluate segmentation under two settings: \textit{Canonical}, where objects are in canonical pose, and \textit{Rotated}, where objects are randomly rotated to emulate arbitrary orientations.


\noindent
\textbf{Results on 3Dcompat-Coarse and 3Dcompat-Fine.} Table~\ref{tab:promptable_seg_combined} reports the quantitative results on the coarse-grained and fine-grained test sets of the 3Dcompat200 dataset. The performance is evaluated using mean Intersection-over-Union (mIoU), computed by averaging part IoUs per object and then across all instances.
Our method consistently outperforms all baselines, achieving an average improvement of $25.55\%$ over the second-best approach, Find3D. Compared to 2D rendering-based methods (e.g., PartSLIP++~\cite{zhou2023partslip++}, PointCLIPV2~\cite{zhu2023pointclip}), our model not only enables significantly faster inference ($0.9$ seconds per shape vs. $2.5$ minutes for PartSLIP++) due to its feedforward mechanism, but also achieves higher segmentation accuracy. Furthermore, compared to the 3D feedforward method Find3D~\cite{ma2025find}, for which we report results from both the original author-released model and a re-trained variant on our curated dataset for fairness, our method demonstrates consistent improvements across all test settings. On the coarse-grained dataset, our model achieves $8\%-11\%$ absolute mIoU improvement over Find3D across both canonical and rotated poses. On the fine-grained dataset, we observe $4\%-7\%$ gains under the same settings. These improvements highlight the effectiveness of our canonical-space-guided learning strategy in enhancing semantic understanding, under pose variations and different prompt forms.

\begin{table*}[t]
\centering
\caption{Quantitative results of the ablation study. All values are mIoU scores computed over parts from all instances across all categories.
}
\label{tab:ablation}
\resizebox{\textwidth}{!}{%
\scriptsize
\begin{tabular}{ccccc|cccc}
\toprule
\multirow{2}{*}{Variant} & \multirow{2}{*}{\begin{tabular}{c}Hard-Negative\\Sampling\end{tabular}} & \multirow{2}{*}{\begin{tabular}{c}Canon. Map\\Anchoring\end{tabular}} & \multirow{2}{*}{\begin{tabular}{c}Cross-Cat.\\Canon. (Data)\end{tabular}} & \multirow{2}{*}{\begin{tabular}{c}Canon. Box\\Calibration\end{tabular}} & \multicolumn{2}{c}{Canonical} & \multicolumn{2}{c}{Rotated} \\
\cmidrule(lr){6-7} \cmidrule(lr){8-9}
 & & & & & $\{\text{Part}\}$ of $\{\text{Obj.}\}$ & $\{\text{Part}\}$ & $\{\text{Part}\}$ of $\{\text{Obj.}\}$ & $\{\text{Part}\}$ \\
\midrule
A & & & &  & 36.89 & 45.00 & 38.14 & 46.02 \\
B & \checkmark  & & & &  38.30 & 48.30 & 39.28 & 48.76 \\
C & \checkmark  & \checkmark & & & 41.71 & 52.01 & 42.48 & 52.57 \\
D & \checkmark  & \checkmark & \checkmark & & 43.34 & 54.15 & 42.63 & 53.70 \\
Full Model & \checkmark  & \checkmark & \checkmark & \checkmark & \textbf{47.51} & \textbf{54.52} & \textbf{47.74} & \textbf{54.55} \\
\bottomrule
\end{tabular}%
}
\begin{tablenotes}
\footnotesize
\item * All settings are trained on intra-category canonicalized shapes.
\item * `Cross-Cat. Canon (Data)' denotes processing the dataset via our cross-category canonicalization pipeline.
\end{tablenotes}
\vspace{-5mm}
\end{table*}
\noindent
\textbf{Results on ShapeNet-Part and PartNet-E.}
Table~\ref{tab:promptable_seg_combined} also reports experimental results on the ShapeNet-Part and PartNet-E datasets. Our method achieves state-of-the-art (SOTA) performance on both benchmarks, with average improvements of 29.89$\%$ on ShapeNet-Part and 5.01$\%$ on PartNet-E over the best-performing baselines. Notably, most methods perform moderately on PartNet-E, as its annotations focus on object materials and fine-grained details—features that are typically absent in other 3D semantic datasets. Interestingly, PartSLIP++, as a 2D rendering-based method, achieves relatively better performance under the single-word prompt setting. This is likely attributed to its backbone model GLIP, which was pre-trained on approximately 27 million image-text pairs and further fine-tuned on the canonical data of the PartNet-E dataset. Though 2D rendering–based methods are limited by speed and multi-view inconsistency, these observations underscore the need to better mine structured semantic priors from large 2D datasets, while accounting their noisy labels and domain gaps, and to efficiently leverage this knowledge alongside limited but well-annotated 3D segmentation data.

\noindent\textbf{Qualitative Comparisons.}
Figure~\ref{fig:comparison} presents a qualitative comparison with Find3D and 2D rendering-based baselines across four scenarios.
For geometry-similar but semantically different parts (left), existing methods often confuse strings and hooks with the object body due to their similar elongated shapes, whereas our model accurately localizes each target part.
On noise-prone objects, where small or thin parts are easily distracted by nearby clutter, baselines either miss the part or predict oversized regions, while our predictions remain tight and stable.
For cross-category semantic synonyms such as ``handle'' appearing on different object categories, prior methods struggle to transfer across shapes and frequently drift to nearby regions, whereas our approach produces consistent segments aligned with the intended semantic prompt.
Finally, under arbitrary pose variations, competing methods are sensitive to pose changes, whereas our framework maintains robust localization across poses, thanks to the canonical-space guidance.



\subsection{Ablation Study and Analysis}
\noindent\textbf{Ablation Study.}
We investigate the contribution of each key component by evaluating variants~A--D and the full model. Results are summarized in Table~\ref{tab:ablation}, where variant~A serves as the baseline.
Hard-Negative Sampling (variant~B) modestly improves robustness of the contrastive alignment, while Canonical Map Anchoring loss (variant~C) brings a larger gain by enforcing pose-robust, canonical-space–aware part features.
Although variants~A--C are all trained on intra-category canonical instances, further applying our cross-category canonicalization pipeline (variant~D) on training data consistently boosts performance across all test settings, highlighting the benefit of cross-category canonical supervision.
Finally, adding the Canonical Box Calibration loss yields the best overall results by regularizing the extracted features via constraining the per-part canonical bounding boxes predicted by the canonical branch. Qualitative analysis of different variants is provided in the supplementary material.

\begin{figure}[t]
\centering
  \includegraphics[width=0.99\linewidth]{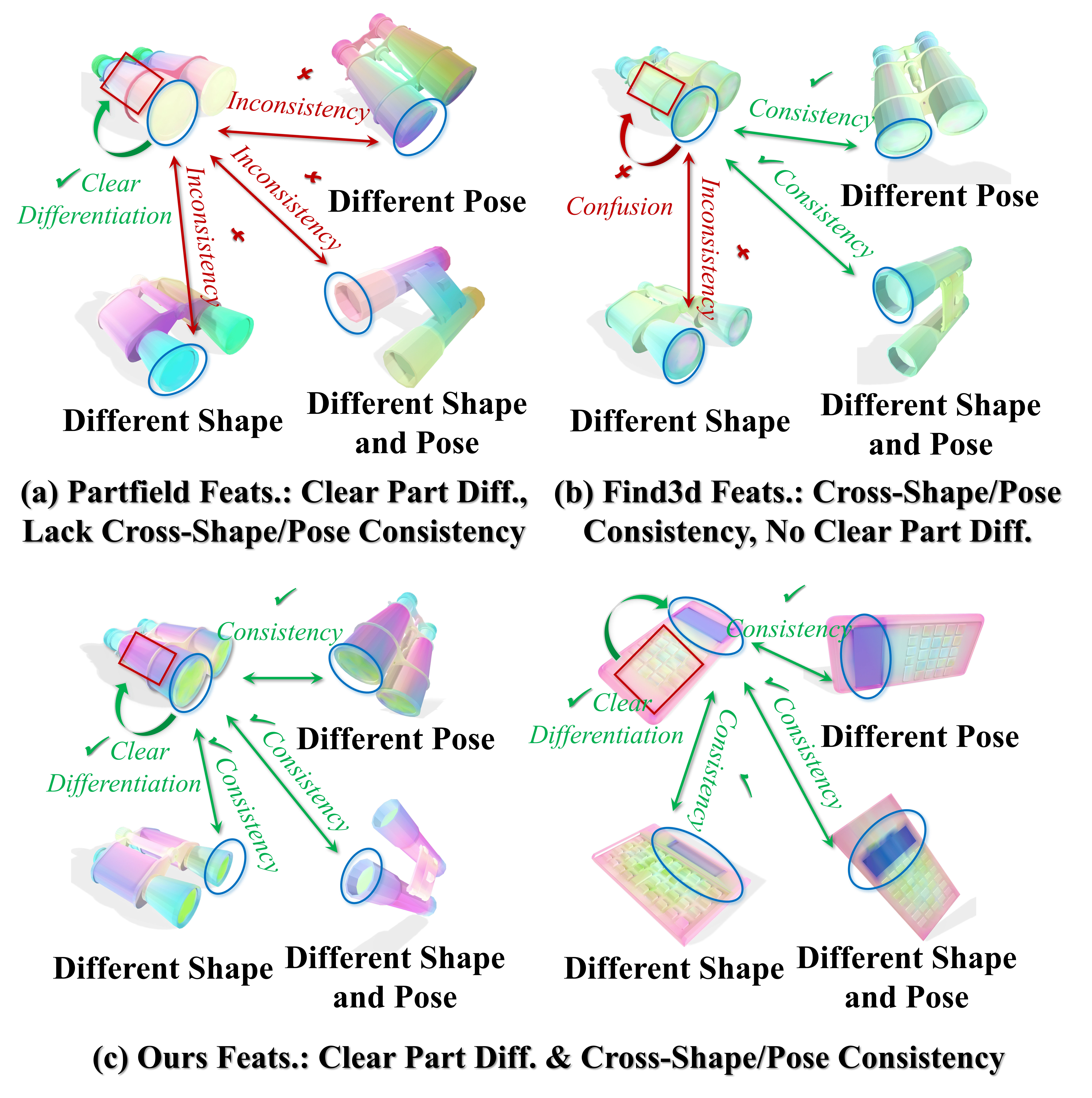} 
\vspace{-3mm}
\caption{Shape feature analysis. 
(a) PartField~\cite{liu2025partfield} yields clear part partitions but lacks cross-shape and cross-pose consistency. 
(b) Find3D~\cite{ma2025find} improves consistency but produces blurry part boundaries. 
(c) Our method achieves both clear part separation and consistent features across shapes and poses.}
  \label{fig:feats}
  \vspace{-0.55cm}
\end{figure}

\noindent\textbf{3D Shape Feature Analysis.}
To better understand the learned representations, we visualize point-wise features from different methods in Fig.~\ref{fig:feats}.
The class-agnostic decomposition method PartField~\cite{liu2025partfield} produces clear part partitions, but its features are not consistent across shapes or poses.
Find3D~\cite{ma2025find} shows stronger cross-shape and cross-pose consistency, yet the features are still not fully aligned across shapes. Moreover, the features from Find3D within a single object lack sharp part boundaries, and points from adjacent parts are often mixed.
In contrast, our framework yields features that are both semantically aligned and structurally well separated: different parts exhibit clear boundaries, and corresponding parts across shapes and poses share highly similar feature patterns.
This behavior stems from training on semantic part segmentation together with our canonical space regularization, which explicitly encourages the network to align geometry with semantics in a shared canonical space. As a result, the learned point-wise features exhibit cross-shape and cross-pose consistency together with fine-grained part distinctions, suggesting that our architecture has the potential to serve as a strong backbone for 3D shape feature learning beyond the specific semantic segmentation task.

\section{Conclusions}
\label{sec:conclusion}
\vspace{-2mm}
We revisited open-world promptable 3D segmentation and identified canonical space as the missing representational variable. While our formulation instantiated this idea through a latent canonical reference frame, we believe its implications extend beyond promptable segmentation.
Once canonicality is a learnable object, canonical space perception could enable richer tasks: compositional 3D query answering, cross-modal grounding across CAD/video domains, and next-stage 3D agents that operate in canonical space before acting in Euclidean space. We view \methodName{} not simply as a system, but as a step toward a more principled 3D understanding stack where the canonical reference is the first-class representational layer.

{
    \small
    \bibliographystyle{ieeenat_fullname}
    \bibliography{main}

@String(TOG= {ACM Trans. Graph.})

@String(Science = {Science})

@String(TOG   = {ACM TOG})

@article{Roger1971mental,
author = {Roger N. Shepard  and Jacqueline Metzler },
title = {Mental Rotation of Three-Dimensional Objects},
journal = {Science},
volume = {171},
number = {3972},
pages = {701-703},
year = {1971}}

@article{chang2015shapenet,
  title={Shapenet: An information-rich 3d model repository},
  author={Chang, Angel X and Funkhouser, Thomas and Guibas, Leonidas and Hanrahan, Pat and Huang, Qixing and Li, Zimo and Savarese, Silvio and Savva, Manolis and Song, Shuran and Su, Hao and others},
  journal={arXiv preprint arXiv:1512.03012},
  year={2015}
}

@inproceedings{wang2019normalized,
  title={Normalized object coordinate space for category-level 6d object pose and size estimation},
  author={Wang, He and Sridhar, Srinath and Huang, Jingwei and Valentin, Julien and Song, Shuran and Guibas, Leonidas J},
  booktitle={Proceedings of the IEEE/CVF conference on computer vision and pattern recognition},
  pages={2642--2651},
  year={2019}
}

@inproceedings{wu2023omniobject3d,
  title={Omniobject3d: Large-vocabulary 3d object dataset for realistic perception, reconstruction and generation},
  author={Wu, Tong and Zhang, Jiarui and Fu, Xiao and Wang, Yuxin and Ren, Jiawei and Pan, Liang and Wu, Wayne and Yang, Lei and Wang, Jiaqi and Qian, Chen and others},
  booktitle={Proceedings of the IEEE/CVF Conference on Computer Vision and Pattern Recognition},
  pages={803--814},
  year={2023}
}

@inproceedings{radford2021learning,
  title={Learning transferable visual models from natural language supervision},
  author={Radford, Alec and Kim, Jong Wook and Hallacy, Chris and Ramesh, Aditya and Goh, Gabriel and Agarwal, Sandhini and Sastry, Girish and Askell, Amanda and Mishkin, Pamela and Clark, Jack and others},
  booktitle={International conference on machine learning},
  pages={8748--8763},
  year={2021},
  organization={PmLR}
}

@inproceedings{liu2023partslip,
  title={Partslip: Low-shot part segmentation for 3d point clouds via pretrained image-language models},
  author={Liu, Minghua and Zhu, Yinhao and Cai, Hong and Han, Shizhong and Ling, Zhan and Porikli, Fatih and Su, Hao},
  booktitle={Proceedings of the IEEE/CVF conference on computer vision and pattern recognition},
  pages={21736--21746},
  year={2023}
}

@inproceedings{ma2025find,
  title={Find any part in 3d},
  author={Ma, Ziqi and Yue, Yisong and Gkioxari, Georgia},
  booktitle={Proceedings of the IEEE/CVF International Conference on Computer Vision},
  pages={7818--7827},
  year={2025}
}

@inproceedings{kirillov2023segment,
  title={Segment anything},
  author={Kirillov, Alexander and Mintun, Eric and Ravi, Nikhila and Mao, Hanzi and Rolland, Chloe and Gustafson, Laura and Xiao, Tete and Whitehead, Spencer and Berg, Alexander C and Lo, Wan-Yen and others},
  booktitle={Proceedings of the IEEE/CVF international conference on computer vision},
  pages={4015--4026},
  year={2023}
}

@article{ravi2024sam,
  title={Sam 2: Segment anything in images and videos},
  author={Ravi, Nikhila and Gabeur, Valentin and Hu, Yuan-Ting and Hu, Ronghang and Ryali, Chaitanya and Ma, Tengyu and Khedr, Haitham and R{\"a}dle, Roman and Rolland, Chloe and Gustafson, Laura and others},
  journal={arXiv preprint arXiv:2408.00714},
  year={2024}
}

@article{tang2024segment,
  title={Segment any mesh: Zero-shot mesh part segmentation via lifting segment anything 2 to 3d},
  author={Tang, George and Zhao, William and Ford, Logan and Benhaim, David and Zhang, Paul},
  journal={arXiv e-prints},
  pages={arXiv--2408},
  year={2024}
}

@article{yang2024sampart3d,
  title={Sampart3d: Segment any part in 3d objects},
  author={Yang, Yunhan and Huang, Yukun and Guo, Yuan-Chen and Lu, Liangjun and Wu, Xiaoyang and Lam, Edmund Y and Cao, Yan-Pei and Liu, Xihui},
  journal={arXiv preprint arXiv:2411.07184},
  year={2024}
}

@article{yang2023sam3d,
  title={Sam3d: Segment anything in 3d scenes},
  author={Yang, Yunhan and Wu, Xiaoyang and He, Tong and Zhao, Hengshuang and Liu, Xihui},
  journal={arXiv preprint arXiv:2306.03908},
  year={2023}
}

@inproceedings{liu2025partfield,
  title={Partfield: Learning 3d feature fields for part segmentation and beyond},
  author={Liu, Minghua and Uy, Mikaela Angelina and Xiang, Donglai and Su, Hao and Fidler, Sanja and Sharp, Nicholas and Gao, Jun},
  booktitle={Proceedings of the IEEE/CVF International Conference on Computer Vision},
  pages={9704--9715},
  year={2025}
}

@article{deng2025geosam2,
  title={GeoSAM2: Unleashing the Power of SAM2 for 3D Part Segmentation},
  author={Deng, Ken and Yang, Yunhan and Sun, Jingxiang and Liu, Xihui and Liu, Yebin and Liang, Ding and Cao, Yan-Pei},
  journal={arXiv preprint arXiv:2508.14036},
  year={2025}
}

@article{zhu2025partsam,
  title={PartSAM: A Scalable Promptable Part Segmentation Model Trained on Native 3D Data},
  author={Zhu, Zhe and Wan, Le and Xu, Rui and Zhang, Yiheng and Chen, Honghua and Dou, Zhiyang and Lin, Cheng and Liu, Yuan and Wei, Mingqiang},
  journal={arXiv preprint arXiv:2509.21965},
  year={2025}
}

@article{yan2025x,
  title={X-Part: high fidelity and structure coherent shape decomposition},
  author={Yan, Xinhao and Xu, Jiachen and Li, Yang and Ma, Changfeng and Yang, Yunhan and Wang, Chunshi and Zhao, Zibo and Lai, Zeqiang and Zhao, Yunfei and Chen, Zhuo and others},
  journal={arXiv preprint arXiv:2509.08643},
  year={2025}
}

@article{zhou2023partslip++,
  title={Partslip++: Enhancing low-shot 3d part segmentation via multi-view instance segmentation and maximum likelihood estimation},
  author={Zhou, Yuchen and Gu, Jiayuan and Li, Xuanlin and Liu, Minghua and Fang, Yunhao and Su, Hao},
  journal={arXiv preprint arXiv:2312.03015},
  year={2023}
}

@inproceedings{zhu2023pointclip,
  title={Pointclip v2: Prompting clip and gpt for powerful 3d open-world learning},
  author={Zhu, Xiangyang and Zhang, Renrui and He, Bowei and Guo, Ziyu and Zeng, Ziyao and Qin, Zipeng and Zhang, Shanghang and Gao, Peng},
  booktitle={Proceedings of the IEEE/CVF international conference on computer vision},
  pages={2639--2650},
  year={2023}
}

@inproceedings{umam2024partdistill,
  title={Partdistill: 3d shape part segmentation by vision-language model distillation},
  author={Umam, Ardian and Yang, Cheng-Kun and Chen, Min-Hung and Chuang, Jen-Hui and Lin, Yen-Yu},
  booktitle={Proceedings of the IEEE/CVF Conference on Computer Vision and Pattern Recognition},
  pages={3470--3479},
  year={2024}
}

@inproceedings{abdelreheem2023satr,
  title={Satr: Zero-shot semantic segmentation of 3d shapes},
  author={Abdelreheem, Ahmed and Skorokhodov, Ivan and Ovsjanikov, Maks and Wonka, Peter},
  booktitle={Proceedings of the IEEE/CVF International Conference on Computer Vision},
  pages={15166--15179},
  year={2023}
}

@inproceedings{zhai2023sigmoid,
  title={Sigmoid loss for language image pre-training},
  author={Zhai, Xiaohua and Mustafa, Basil and Kolesnikov, Alexander and Beyer, Lucas},
  booktitle={Proceedings of the IEEE/CVF international conference on computer vision},
  pages={11975--11986},
  year={2023}
}

@inproceedings{jin2025one,
  title={One-shot 3D Object Canonicalization based on Geometric and Semantic Consistency},
  author={Jin, Li and Wang, Yujie and Chen, Wenzheng and Dai, Qiyu and Gao, Qingzhe and Qin, Xueying and Chen, Baoquan},
  booktitle={Proceedings of the Computer Vision and Pattern Recognition Conference},
  pages={16850--16859},
  year={2025}
}

@article{ahmed20243dcompat200,
  title={3DCoMPaT200: Language Grounded Large-Scale 3D Vision Dataset for Compositional Recognition},
  author={Ahmed, Mahmoud and Li, Xiang and Prajapati, Arpit and Elhoseiny, Mohamed},
  journal={Advances in Neural Information Processing Systems},
  volume={37},
  pages={135770--135782},
  year={2024}
}

@inproceedings{wu2024point,
  title={Point transformer v3: Simpler faster stronger},
  author={Wu, Xiaoyang and Jiang, Li and Wang, Peng-Shuai and Liu, Zhijian and Liu, Xihui and Qiao, Yu and Ouyang, Wanli and He, Tong and Zhao, Hengshuang},
  booktitle={Proceedings of the IEEE/CVF conference on computer vision and pattern recognition},
  pages={4840--4851},
  year={2024}
}

@article{li2024craftsman3d,
  title={Craftsman3d: High-fidelity mesh generation with 3d native generation and interactive geometry refiner},
  author={Li, Weiyu and Liu, Jiarui and Yan, Hongyu and Chen, Rui and Liang, Yixun and Chen, Xuelin and Tan, Ping and Long, Xiaoxiao},
  journal={arXiv preprint arXiv:2405.14979},
  year={2024}
}

@article{hunyuan3d2025hunyuan3d,
  title={Hunyuan3D 2.1: From Images to High-Fidelity 3D Assets with Production-Ready PBR Material},
  author={Hunyuan3D, Team and Yang, Shuhui and Yang, Mingxin and Feng, Yifei and Huang, Xin and Zhang, Sheng and He, Zebin and Luo, Di and Liu, Haolin and Zhao, Yunfei and others},
  journal={arXiv preprint arXiv:2506.15442},
  year={2025}
}

@article{takmaz2023openmask3d,
  title={Openmask3d: Open-vocabulary 3d instance segmentation},
  author={Takmaz, Ay{\c{c}}a and Fedele, Elisabetta and Sumner, Robert W and Pollefeys, Marc and Tombari, Federico and Engelmann, Francis},
  journal={arXiv preprint arXiv:2306.13631},
  year={2023}
}

@inproceedings{zhang2025generative,
  title={Generative Hard Example Augmentation for Semantic Point Cloud Segmentation},
  author={Zhang, Qi and Peng, Jibin and Huang, Zhao and Feng, Wei and Lin, Di},
  booktitle={Proceedings of the Computer Vision and Pattern Recognition Conference},
  pages={22205--22214},
  year={2025}
}

@article{ma2022rethinking,
  title={Rethinking network design and local geometry in point cloud: A simple residual MLP framework},
  author={Ma, Xu and Qin, Can and You, Haoxuan and Ran, Haoxi and Fu, Yun},
  journal={arXiv preprint arXiv:2202.07123},
  year={2022}
}

@inproceedings{zhao2021point,
  title={Point transformer},
  author={Zhao, Hengshuang and Jiang, Li and Jia, Jiaya and Torr, Philip HS and Koltun, Vladlen},
  booktitle={Proceedings of the IEEE/CVF international conference on computer vision},
  pages={16259--16268},
  year={2021}
}

@article{girshick2015region,
  title={Region-based convolutional networks for accurate object detection and segmentation},
  author={Girshick, Ross and Donahue, Jeff and Darrell, Trevor and Malik, Jitendra},
  journal={IEEE transactions on pattern analysis and machine intelligence},
  volume={38},
  number={1},
  pages={142--158},
  year={2015},
  publisher={IEEE}
}

@inproceedings{mo2019partnet,
  title={Partnet: A large-scale benchmark for fine-grained and hierarchical part-level 3d object understanding},
  author={Mo, Kaichun and Zhu, Shilin and Chang, Angel X and Yi, Li and Tripathi, Subarna and Guibas, Leonidas J and Su, Hao},
  booktitle={Proceedings of the IEEE/CVF conference on computer vision and pattern recognition},
  pages={909--918},
  year={2019}
}

@article{yi2016scalable,
  title={A scalable active framework for region annotation in 3d shape collections},
  author={Yi, Li and Kim, Vladimir G and Ceylan, Duygu and Shen, I-Chao and Yan, Mengyan and Su, Hao and Lu, Cewu and Huang, Qixing and Sheffer, Alla and Guibas, Leonidas},
  journal={ACM Transactions on Graphics (ToG)},
  volume={35},
  number={6},
  pages={1--12},
  year={2016},
  publisher={ACM New York, NY, USA}
}
}


\end{document}